\documentclass{article}
\usepackage{spconf,amsmath,graphicx}
\usepackage[utf8]{inputenc} 
\usepackage[T1]{fontenc}    
\usepackage{hyperref}       
\usepackage{url}            
\usepackage{booktabs}       
\usepackage{amsfonts}       
\usepackage{nicefrac}       
\usepackage{microtype}      

\usepackage{graphicx}  
\usepackage{chngpage}

\usepackage{amsmath}
\usepackage{mathtools}
\usepackage[utf8]{inputenc}

\usepackage{tabularx}

\usepackage{arydshln}
\usepackage{array,etoolbox}
\preto\tabular{\setcounter{magicrownumbers}{0}}
\newcounter{magicrownumbers}

\usepackage{times}  
\usepackage{helvet}  
\usepackage{courier}  
\usepackage{latexsym}
\usepackage{color}
\usepackage{amsmath}
\usepackage{mathtools}
\usepackage{amssymb}
\usepackage{amsfonts}
\usepackage{breqn}
\usepackage{float}

\usepackage{algorithm}
\usepackage[noend]{algpseudocode}

\usepackage{caption}
\usepackage{subcaption}
\usepackage{paralist}

\newcommand\enc[1]{\textnormal{enc}\left(#1\right)}
\newcommand\dec[1]{\textnormal{dec}\left(#1\right)}
\newcommand\transl[1]{\texttt{transl}\left(#1\right)}
\newcommand\sli{s_{l_i}}

\newcolumntype{L}[1]{>{\raggedright\let\newline\\\arraybackslash\hspace{0pt}}m{#1}}
\newcolumntype{C}[1]{>{\centering\let\newline\\\arraybackslash\hspace{0pt}}m{#1}}
\newcolumntype{R}[1]{>{\raggedleft\let\newline\\\arraybackslash\hspace{0pt}}m{#1}}

\usepackage{url}

\title{FROM UNSUPERVISED MACHINE TRANSLATION TO ADVERSARIAL TEXT GENERATION}
\name{Ahmad Rashid \qquad Alan Do-Omri \qquad Md. Akmal Haidar \qquad Qun Liu \qquad Mehdi Rezagholizadeh}

\address{Huawei Noah's Ark Lab, Montreal Research Centre, Canada \\
{\fontsize{10pt}{10pt}\selectfont  \{ahmad.rashid, alan.do.omri, md.akmal.haidar, qun.liu, mehdi.rezagholizadeh\}@huawei.com}}
\begin{document}
%
\maketitle
\begin{abstract}
We present a self-attention based bilingual adversarial text generator (B-GAN) which can learn to generate text from the encoder representation of an unsupervised neural machine translation system. B-GAN is able to generate a distributed latent space representation which can be paired with an attention based decoder to generate fluent sentences. When trained on an encoder shared between two languages and paired with the appropriate decoder, it can generate sentences in either language. B-GAN is trained using a combination of reconstruction loss for auto-encoder, a cross domain loss for translation and a GAN based adversarial loss for text generation. We demonstrate that B-GAN, trained on monolingual corpora only using multiple losses, generates more fluent sentences compared to monolingual baselines while effectively using half the number of parameters.
\end{abstract}
\section{Introduction}
\label{sec:intro}


Language generation is a vital component of many Natural Language Processing (NLP) applications including dialogue systems, question answering, image captioning and summarization. State of the art systems \cite{radfordlanguagegtpt,guo2017long}, however, can only generate text in one language at a time. Consequently, we train and deploy one model for each language we are interested in. This is specially difficult for edge deployment of machine learning models. Our work presents a bilingual text generation model, B-GAN, which is able to generate text in two languages simultaneously while using the same number of parameters as monolingual baselines.


\begin{figure}[]
 \centering
 \includegraphics[scale=0.2]{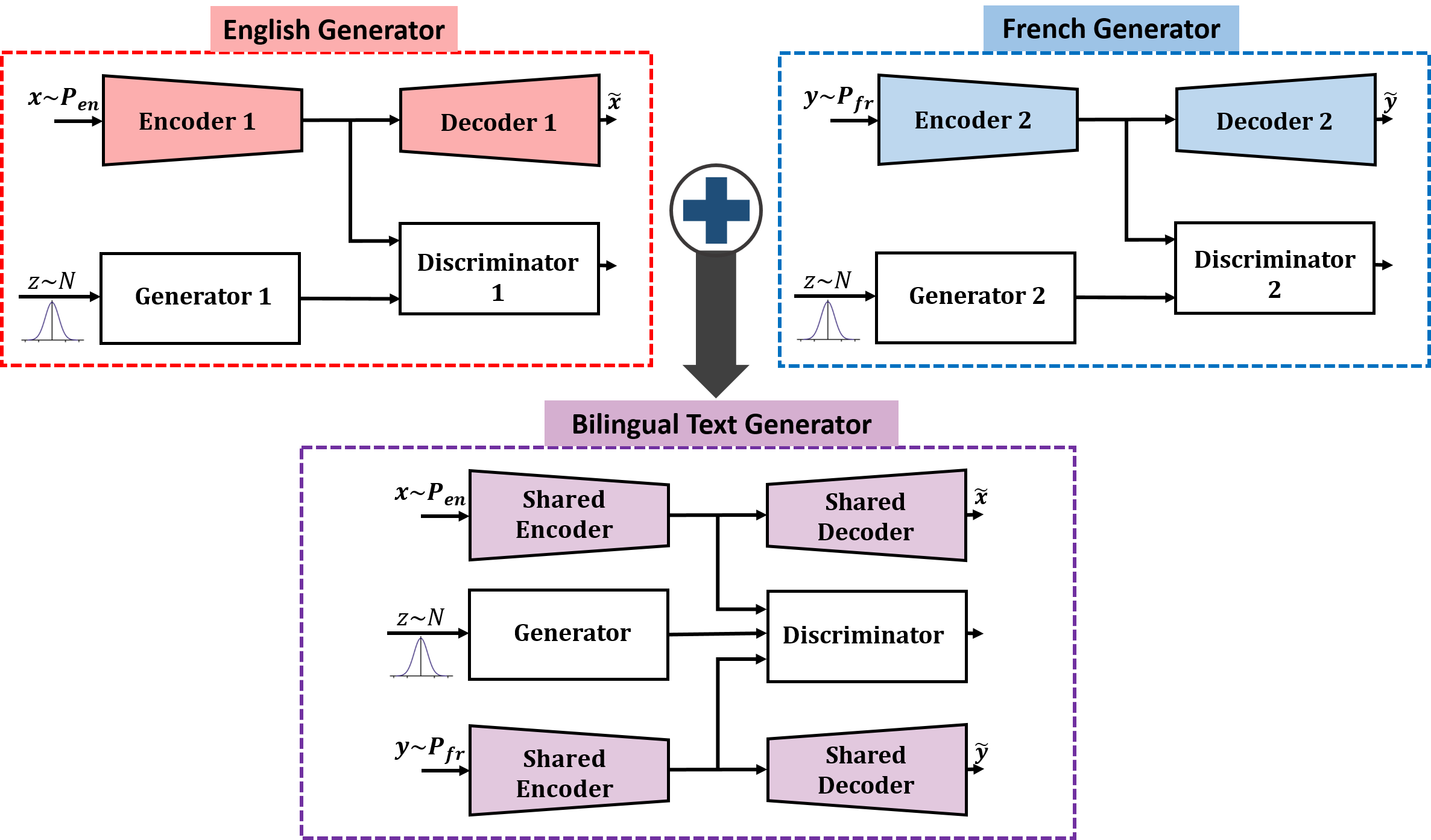}
 \caption{Traditional multilingual text generation vs. our bilingual text generation technique with a shared autoencoder}\label{fig:Intro}
 \end{figure}

Neural text generation predominantly employs autoregressive methods based on Maximum Likelihood Estimation (MLE). They use teacher forcing for their training which might give rise to shortsightedness and exposure bias~\cite{scheduled_sampling}. A number of solutions have been proposed including scheduled sampling~\cite{scheduled_sampling} and Professor forcing~\cite{lamb2016professor}. One alternative is Generative Adversarial Networks (GANs)~\cite{goodfellow2014generative} a generative model for text~\cite{gulrajani2017improved}, which has shown impressive results in image generation. 
Adversarial neural text generation, however, faces a few challenges including the discrete nature of text, quality vs. diversity of the sentences, and mode collapse. 

Our work combines adversarial-based text generation and autoregressive models~\cite{kim2017adversarially, gagnon2018salsa, Akmal2019a}. Building on the idea of unsupervised machine translation in~\cite{FAE,lample2018phrase}, we use the concept of shared encoders between two languages and multi-lingual embeddings to learn the aligned latent representation of two languages and a GAN which can sample from this latent space to generate text (see Fig.~\ref{fig:Intro}). \cite{rashid2019bilingual} combines machine translation with GANs to explore parallel text generation but they can only generate short sentences. However, inspired by ~\cite{gagnon2018salsa}, we use self-attention architectures along with unsupervised machine translation and adversarial generation to generate longer, fluent sentences. We propose B-GAN, an agent capable of deriving a shared latent space between two languages, and then generating from this space in either language. In summary:
\begin{itemize}
    \item We propose a single system, B-GAN, trained on monolingual data that can generate text in two languages simultaneously while effectively using half the number of parameters per language.
    \item B-GAN learns to match the encoder representation of an unsupervised machine translation system. 
    \item B-GAN generates more fluent text compared to monolingual baselines on quantitative and qualitative evaluation. 
\end{itemize}

\section{Related Work}
In this section we discuss some of the recent developments in sequence to sequence learning~\cite{sutskever2014sequence} and adversarial text generation which are relevant to our work.

\textbf {Unsupervised NMT} A few recent works \cite{FAE,UNdreaMT,lample2018phrase} have pushed the frontier of machine translation systems trained on only monolingual corpora. The common principles of such systems include learning a language model, encoding sentences from different languages into a shared latent representation and using back-translation \cite{back_translation} to provide pseudo supervision. 

\textbf{ARAE} Applying GAN to text generation is challenging due to the discrete nature of text. Consequently, back-propagation is not feasible for discrete outputs and computing gradients of the output of decoder requires approximations of the argmax operator. A latent code-based solution for this problem was proposed in ~\cite{kim2017adversarially}, where instead of learning to generate reconstructed text the generator learns the latent representation of the encoder. Once trained, a decoder is used to convert the sampled representations into text. This model, however, can only generate text in one language.

\textbf{Spectral Normalization}~\cite{miyato2018spectral} is a weight normalization method proposed to bound the Lipshitz norm of neural networks by normalizing the spectral norm of layer weight matrices. This is in contrast to local regularization used in WGAN-GP \cite{gulrajani2017improved}. The authors show that spectral normalization of the discriminator weights stabilizes GAN training and produces more diverse outputs for image generation.

\textbf{Self-Attention}~\cite{parikh2016decomposable}, an alternative to LSTMs \cite{LSTM}, is a method to learn a representation of sequences. It has been shown to encode long term dependencies well, is non-autoregressive and highly parallelizable. Transformer~\cite{vaswani2017attention}  extended the use of self attention mechanism to sequence transduction applications such as machine translation and is the basis of many state of the art systems~\cite{devlin2018bert}. \cite{sagan} applied self attention along with spectral normalization to the task of image generation using GANs.  \cite{gagnon2018salsa} extended this technique to text generation. B-GAN is built with self-attention can generate text in multiple languages unlike the previous works. 

\section{Methodology}
\begin{figure}[htb!]
 \centering
 \includegraphics[width=\linewidth]{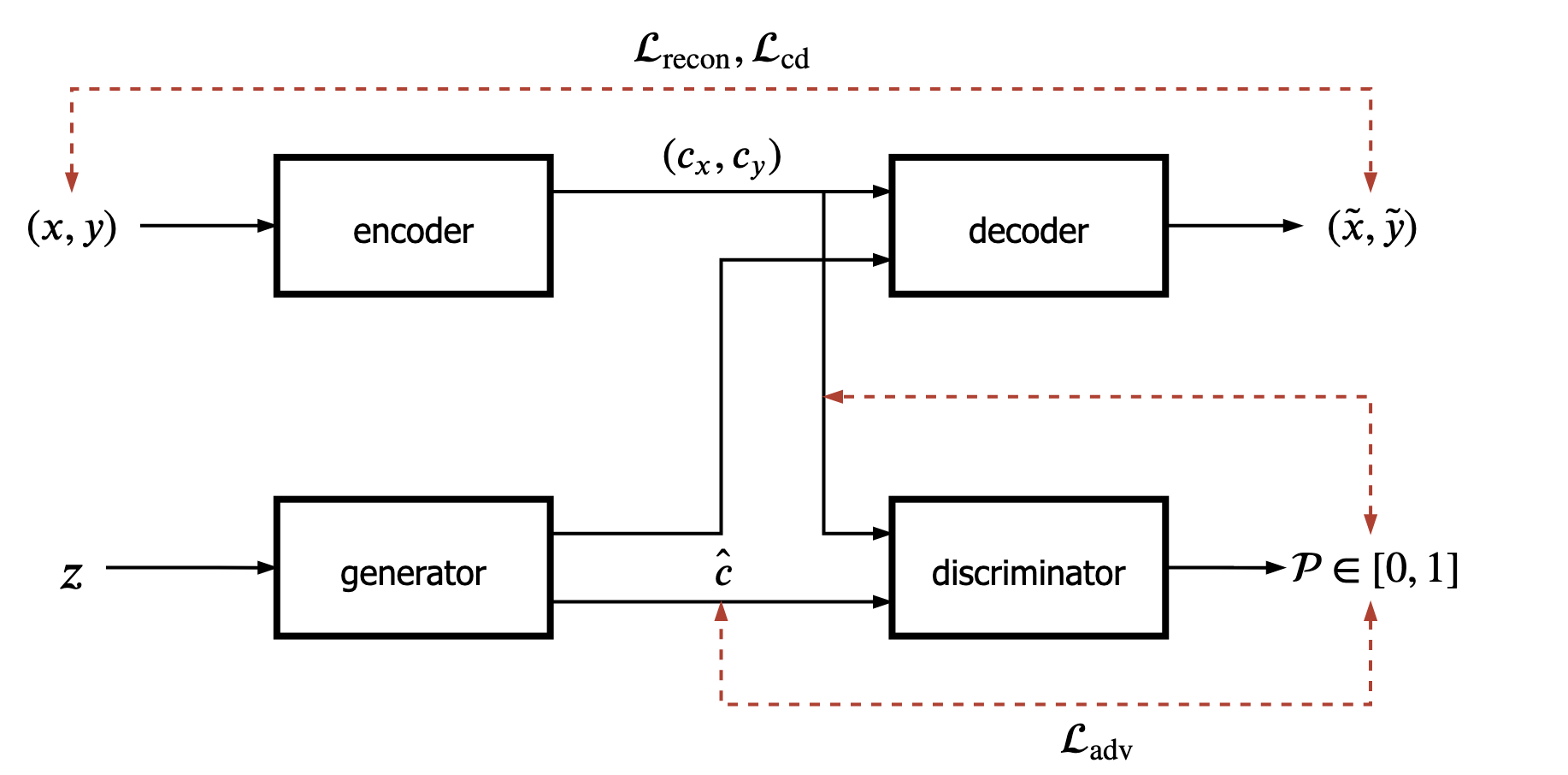}
 \caption{The complete architecture of B-GAN}\label{fig:BGAN}
 \end{figure}

B-GAN comprises of two main components: a translation unit trained on cross-entropy loss for reconstruction ($\mathcal{L}_{recon}$) and translation ($\mathcal{L}_{cd}$) and a text generation unit trained on adversarial loss ($\mathcal{L}_{adv}$). The complete architecture is illustrated in Figure~\ref{fig:BGAN}. 

\subsection{Translation Unit}
\label{translation}
The translation system is a Transformer based sequence-to-sequence model similar to~\cite{lample2018phrase}. The key components of this system are a denoising auto-encoder for both language 1 and language 2, on the fly back translation~\cite{UNdreaMT} and aligned latent representations. The latent representations or the code for both languages are aligned by sharing the encoder weights for the two languages, using joint word embeddings and sub-word tokens~\cite{BPE}. Sub-word tokens have a high overlap for related language pairs such as English and French. 

\par We apply the token wise cross-entropy loss to train our model. Let $\sli$ be a sentence in language $i$ with $i \in \{1,2\}$. 
We denote the encoding of sentence $\sli$ by $\enc{\sli}$ and similarly, the deocoding of code $x$ (typically an output of the encoder) into language $l_i$ as $\dec{x, l_i}$. 
\par The system is trained with two losses aimed to allow the encoder-decoder pair to reconstruct inputs (reconstruction loss) and to translate correctly (cross-domain loss).

\textbf{Reconstruction Loss}, which, is the standard auto-encoder loss which aims to reconstruct the input:
\begin{equation} 
\small
\mathcal{L}_{\textnormal{recon}} = \Delta\left(\sli, \overbrace{\dec{\enc{\sli}, l_i}}^{\mathclap{\hat{s}_{l_i}} \quad \coloneqq}\right)
\label{eq:recon}
\end{equation}

\textbf{Cross-Domain Loss}, which aims to allow translation of inputs. It is similar to back-translation \cite{back_translation}. For this loss, denote by $\transl{\sli}$ the translation of sentence $\sli$ from language $i$ to language $3-i$. 

\begin{equation}
\small
\mathcal{L}_{\textnormal{cd}} = \Delta\left(\sli, \underbrace{\dec{\enc{\transl{\sli}},l_i}}_{\mathclap{\tilde{s}_{l_i} \; \coloneqq}} \right) \label{eq:cd}
\end{equation}

\subsection{Bilingual Text Generation Unit }

The text generation system is based on a Generative Adversarial Network (GAN)~\cite{goodfellow2014generative}. The generator learns to match the distribution of the latent space of the encoder \cite{ARAE}. The discriminator is fed encoded sentences and generated latent representations and learns to distinguish between the two. The learning process is a two player minimax game between the generator and the discriminator. The discriminator D and generator G are parameterized using neural networks. The latent distribution is $P(c_x)$,$P(c_y)$ where $c_x = Enc(x)$, $c_y = Enc(y)$ is obtained by applying the Encoder, $Enc$, to the sentences $x$ and $y$ (see Figure~\ref{fig:BGAN}) from language $1$ and $2$ respectively. Since we share the latent space for the two languages we assume that if sentence $x_i$ and sentence $y_i$ are translations of each other their latent representations $Enc(x_i)$ and $Enc(y_i)$ are also close under some distant measure. 

\textbf {Adversarial Loss} We employ the hinge version of the adversarial loss to train our generative model. For sentences $x$ this would be:
\begin{equation}
\begin{aligned}
    \mathcal{L_D} = \mathbb{E}_{x\sim{Pdata}}[\min(0, -1 + D(Enc(x)))] \\
    + \mathbb{E}_{z\sim{P(z)}}[\min(0, -1 - D(G(z)))]  
\end{aligned}
\end{equation}

\begin{equation}
\begin{aligned}
    \mathcal{L_G} = - \mathbb{E}_{z\sim{P(z)}}[D(G(z)))]
\end{aligned}
\end{equation}
where $L_D$ and $L_G$ are the discriminator and generator losses respectively.

The architecture of our generator and discriminator is as described in \cite{gagnon2018salsa}. Typically latent space based generators~\cite{ARAE,cifka} match the last hidden state of an LSTM however, our system learns to generate a sequences that can match the distribution of the encoder of a transformer.

\textbf{Training} In each iteration we train, in order, the denoising auto-encoder on English, the same denoising auto-encoder on French, back translation from English to French to English and viceversa, one discriminator update and one generator update. The discriminator is trained on $c_x$ during odd iterations and on $c_y$ during even iterations. 
$Enc$ is a deterministic function and maps data to discrete points inside a continuous space. However the GAN generator produces a continuous distribution so we add Gaussian noise to the encoder representation for better distribution matching. We also apply spherical normalization to the output of the encoder to aid training.





\section{Experiments}

We used Multi30k~\cite{multi30k} and WMT monolingual News Crawl datasets \footnote{http://www.statmt.org/wmt14/translation-task.html} for our experiments. Multi30k consists of 29k images and their captions. We only use the French and English paired captions as the training set and the provided validation set and Flickr 2016 test set. We split the 29k captions into non-overlapping halves~\cite{FAE}. We use News Crawl 2007 to 2010 for both English and French and sample one million sentences each. The validation set is newstest 2013 and the test set is newstest 2014. The test and validation sets are 1k each for Multi30k and 3k each for News Crawl. We tokenize the sentences using the Moses~\footnote{https://github.com/moses-smt/mosesdecoder} tokenizer and combine the English and French News Crawl corpora to learn 60k Byte-Pair Encoded (BPE) \cite{BPE} tokens. We train cross-lingual word embeddings using FastText~\cite{fasttext}. The News Crawl trained embeddings and dictionary are used for Multi30k as well. We remove sentences longer than $T$=35 tokens on Multi30k (0.05\%)  and $T$=50 tokens on News Crawl (7.25\%) where $T$ is the maximum sequence length we can generate.

We present the specification of all our models in Table~\ref{table:models}. We compare B-GAN against three baselines. The baseline ARAE model, which is our implementation of \cite{ARAE}, ARAE$_{Conv}$, where we add 1D convolutions to the GAN and ARAE$_{SA}$, which is based on self-attention \cite{gagnon2018salsa}. 

\begin{table}[!thb]  \small
  \centering
  \begin{tabular}{|l|l|l|l|l|}
  \multicolumn{5}{c}{\textbf{Models}} \\
    \hline
           &B-GAN &ARAE$_{SA}$ &ARAE$_{Conv}$ &ARAE    \\
    \hline
    \textit{Enc-Dec}&SA &SA &LSTM &LSTM \\
    \textit{\hspace*{2pt} Attention} &Yes &Yes &No &No \\
    \textit{\hspace*{2pt} Layers} &4 &4  &2    &1  \\
    \textit{\hspace*{2pt} Back-Trans} &Yes  &No &No   &No  \\
    \hline
    \textit{Gen-Disc} &SA &SA &Conv &MLP \\
    \textit{\hspace*{2pt} Sub-Layers} &2 &2 &2 &2   \\
    \textit{\hspace*{2pt} Spectral Norm} &Yes &Yes &No &No  \\  
    \textit{\hspace*{2pt} Loss} &Hinge &Hinge &WGAN &WGAN \\
    \hline
    \textit{Embedding} &512 &512 &512 &512 \\
    \hline
    \textit{Bilingual} &Yes &No &No &No \\
    \hline
  \end{tabular}
  \caption{Model details for our system, B-GAN, and the three ARAE based systems}
  \label{table:models}
\end{table}

\begin{table*}[h]
  
  \small
  
  \centering
  \begin{tabular}{|l|l|l|l|l|l|l|l|l|}
  \multicolumn{9}{c}{\textbf{Multi30k}} \\ \hline
     &  \multicolumn{4}{c|}{\textbf{English}}&\multicolumn{4}{|c|}{\textbf{French}} \\ \hline
           &B-GAN & ARAE$_{SA}$&ARAE$_{Conv}$&ARAE&B-GAN & ARAE$_{SA}$&ARAE$_{Conv}$& ARAE    \\
           
    \hline
    \textit{B-2} &88.76 & 80.05 &78.00 &92.81    &88.42 &81.93 &90.35 &93.79\\
    \textit{B-3} &75.59 & 60.44 &55.16 &78.57    &76.56 &66.53 &76.56 &82.05\\
    \textit{B-4} &60.20 & 41.84 &34.30 &60.22    &62.42 &50.11 &59.39 &65.37\\
    \textit{B-5} &\textbf{44.16} & 26.90 &19.30 &42.81  &\textbf{47.32} &34.49 &41.69 &47.60\\
    \hline
    \multicolumn{9}{c}{\textbf{News Crawl}} \\ \hline
    \textit{B-2} &76.72 &79.05 &70.54 &76.22   &75.39 &75.23 &58.53 &74.85 \\
    \textit{B-3} &52.08 &53.67 &41.18 &46.04   &51.76 &51.10 &29.72 &46.06 \\
    \textit{B-4} &30.22 &29.70 &18.10 &21.03   &30.43 &30.43 &12.10 &23.42 \\
    \textit{B-5} &\textbf{15.76} &14.56 &7.36  &8.51    &16.24 &\textbf{16.34} &4.23  &10.60 \\
    \hline
  \end{tabular}
  \caption{BLEU scores for Text Generation using 10000 and 100000 generated sentences for the Multi30k and News Crawl datasets respectively (higher is better).}
  \label{bleu-n}
\end{table*}
\vspace{-.5cm}
\subsection{Quantitative Evaluation Metrics}

\textbf{Corpus-level BLEU} We use the BLEU-N scores to evaluate the generated sentences according to~\cite{papineni2002}. BLEU-N is a measure of the fluency of the generated sentences. We also use \textbf{Perplexity} to evaluate the fluency. The forward perplexity (F-PPL) is calculated by training an RNN language model on real  training  data and evaluated  on  the generated samples. We calculate  the  reverse  perplexity (R-PPL) by  training  an RNN language model (LM) on the synthetic samples and evaluating on the real test data. This gives us a measure of the diversity. 


\subsection{Quantitative Evaluation}
We generated 100k  and 10k sentences for BLEU-score evaluations, for the models trained on News Crawl and Multi30k datasets respectively. The BLEU scores of the generated sentences with respect to the test set are presented in Table~\ref{bleu-n}.The higher BLEU scores demonstrate that the model can generate fluent sentences. We note that our proposed B-GAN model can generate more fluent sentences both in English and French compared to the other ARAE models for the larger News Crawl dataset and as fluent on the smaller Multi30k dataset. 

We generated 100k and 10k sentences for the News Crawl and the Multi30k datasets respectively for perplexity analysis. We use a vocabulary of 7208 words for Multi30k and 10000 words for News Crawl for training the LM on the real data for F-PPL. The LM is trained on the generated sentences for R-PPL. The forward and reverse perplexities are displayed in Table~\ref{ppl}.When the forward perplexities of the generated sentences are lower than real data the generated sentences are not as diverse as the real sentences. Their relative diversity can be compared using R-PPL. 

On the Multi30k dataset, our proposed B-GAN model can generate more fluent and diverse sentences in both languages compared to the other models. For the News Crawl dataset, the B-GAN model generates the most fluent but the least diverse sentences. The B-GAN model loses on diversity on the larger dataset as it trains one model for two languages and has therefore fewer parameters.

\begin{table}[h]
    \small
    \centering
  \begin{tabular}{|l|l|l|l|l|}
  \multicolumn{5}{c}{\textbf{Multi30k}} \\ \hline
     &  \multicolumn{2}{c|}{\textbf{English}}&\multicolumn{2}{|c|}{\textbf{French}} \\ \hline
           &\textit{F-PPL} &\textit{R-PPL}  &\textit{F-PPL} &\textit{R-PPL}  \\
    \hline

      
      \hline
      Real     & 32.3 & - & 22.2 & -\\
      ARAE  & 11.2  & 191.1 & 6.9   & 141.9\\
      ARAE$_{Conv}$  & 21.3  & 174.9 & 8.4  & \textbf{89.0}\\
      ARAE$_{SA}$ & 15.9  & 97.0  & 11.9 & 178.1\\
      B-GAN & \textbf{7.5} & \textbf{92.1} & \textbf{6.1}  & 90.0 \\
       \hline

\multicolumn{5}{c}{\textbf{News Crawl}} \\ \hline
      Real     & 132.4 & - & 81.0 & -\\
      ARAE  & 55.6 & 533.9 & 51.7 & 330.0\\
      ARAE$_{Conv}$  & 64.2  & \textbf{325.9} &  137.5 & \textbf{207.7}\\
      ARAE$_{SA}$ & 27.2  & 383.6  & 25.2 & 225.9\\
      B-GAN  & \textbf{26.4}  &971.4 &  \textbf{18.4} & 746.4\\
      
      \hline

  \end{tabular}
  \caption{Forward (F) and Reverse (R) word perplexity (PPL) results on multi30k and News Crawl datasets respectively. Lower is better.
  }
  \label{ppl}
\end{table}

\vspace{-.5cm}
\subsection{Human Evaluation}
We present human evaluation of the generated sentences trained using the News Crawl and the Multi30k datasets in Table~\ref{human_eval}. For our human evaluation experiment, we used 20 random generated sentences from each model. The task was given to a group of 8 people, 4 native in French and 4 in English. Participants were asked to rate the fluency of sentences on a scale of 1 to 5 where the score 1 corresponds to gibberish, 3 to understandable but ungrammatical, and 5 to naturally constructed and understandable sentences~\cite{cifka}. B-GAN generates the most fluent sentences. Table \ref{sen-News Crawl} presents a few sampled sentences. 

\begin{table}[!htb]
    \small
    \centering
  \begin {tabular}{ cc }
  \begin{tabular}{|l|c|c|}
  
  \multicolumn{3}{c}{\textbf{Multi30k}} \\ \hline
  &  \multicolumn{2}{c|}{\textbf{Fluency}}\\
    Models &  \textbf{(EN)}&\textbf{(FR)}  \\ \hline

      Real   & 4.87&4.79  \\
      ARAE & 3.86  & 3.05\\
      ARAE$_{Conv}$ & 3.56&   3.11\\
      ARAE$_{SA}$ & 4.05& 3.53\\
      B-GAN & \textbf{4.76}& \textbf{3.85}\\
      \hline
      \end{tabular}
    \begin{tabular} {|l|c|c|}
    \multicolumn{3}{c}{\textbf{News Crawl}} \\ \hline
    &  \multicolumn{2}{c|}{\textbf{Fluency}}\\
    Models &  \textbf{(EN)}&\textbf{(FR)}  \\ \hline
    
      Real     & 4.89 & 4.81 \\
      ARAE & 2.71 & 1.94 \\
      ARAE$_{Conv}$& 3.05 &  1.56  \\
       ARAE$_{SA}$& 3.73 & \textbf{3.33} \\
       B-GAN & \textbf{4.48} & \textbf{3.33}  \\
      \hline

  \end{tabular} \\ 
  \end{tabular}
  \caption{Human evaluation on the generated sentences by B-GAN using the Europarl and the Multi30k dataset.}
  
  \label{human_eval}
\end{table}

\begin{table}[!htb]
    \small
    \centering
  \begin{tabular}{|l|l|}
    \hline
     \text{1} & CNN 's John McCain and his wife , Maria , were \\ & in the midst of a three-day visit to the country.\\
     \text{2} & The second attack occurred as a result of the attack, \\& and the other two were still being investigated for \\& the death of a British soldier who was also killed in the \\& southern city of Basra .\\
     \hline
     \text{1} & Consistoire : une étude de faisabilité des travaux d'études \\& sur les coûts de l' énergie et des coûts de production \\& des matières premières est en cours d' achèvement.\\
     \text{2} & Ce dernier , qui a été retenu à la base de l' équipe , \\& a été légèrement. \\
     \hline
  \end{tabular}
  \caption{English (top) and French (bottom) sentences generated by B-GAN trained on News Crawl}
  \label{sen-News Crawl}
\end{table}

\vspace{-.5cm}
\section{Conclusion}
\par The motivation of B-GAN is a) to improve the quality of adversarial text generation and b) generate two, or potentially more, languages by training on the space of unsupervised machine translation. We combine cross-entropy loss for text reconstruction and translation respectively to adversarial loss and generate longer and more fluent sentences compared to our baseline. We use self-attention not only for machine translation but also for our GAN.  Our experiments show that our single model which avoids the need of multiple monolingual models yields more fluent sentences in both languages.


\ninept
\bibliographystyle{IEEEbib}
\bibliography{strings,refs}

\end{document}